\documentclass{article}
\usepackage{ijcai20}
\usepackage{times}
\usepackage{soul}
\usepackage{url}
\usepackage[hidelinks]{hyperref}
\usepackage[utf8]{inputenc}
\usepackage[small]{caption}
\usepackage{graphicx}
\usepackage{amsmath}
\usepackage{amsthm}
\usepackage{amsfonts,amssymb}
\usepackage{booktabs}
\usepackage{algorithm}
\usepackage{algorithmic}
\usepackage{colonequals}
\usepackage{xspace}
\usepackage{xcolor}
\usepackage{colortbl,tabularx}
\usepackage{arydshln}
\usepackage{nicefrac}
\usepackage{tikz}
\usepackage{multirow}
\usepackage{url}
\usepackage{pgfplots}
\usepackage{todonotes}
\usepackage{acronym}
\usepackage{mdframed}
\usepackage{subfig}
\usepackage[symbol]{footmisc}
\usetikzlibrary{patterns,arrows,arrows.meta,calc,shapes,shadows,decorations.pathmorphing,decorations.pathreplacing,automata,shapes.multipart,positioning,shapes.geometric,fit,circuits,trees,shapes.gates.logic.US,fit}
\usetikzlibrary{backgrounds,scopes}

\newtheorem{definition}{Definition}

\newtheorem{example}{Example}

\newcommand{\ie}{i.\,e.,\@\xspace}

\newcommand{\tool}[1]{\textrm{#1}\xspace}

\newcommand{\p}{\ensuremath{\mathbb{P}}}

\newcommand{\reachPropSymbol}{\varphi}
\newcommand{\ltlformula}{\psi}

\newcommand{\rew}{\ensuremath{r}}

\newcommand{\R}{\mathbb{R}}

\newcommand{\Reals}{\ensuremath{\mathbb{R}}\xspace}    

\newcommand{\Ireal}{[0,\, 1]\subseteq\mathbb{R}}  
\newcommand{\Ex}{\ensuremath{\mathbb{E}}\xspace}        

\newcommand{\Distr}{\mathit{Distr}}

\newcommand{\distDom}{X}

\newcommand{\distFunc}{\mu}
\newcommand{\distDomElem}{x}
\DeclareMathOperator{\supp}{supp}

\newcommand{\Always}{\Box\,}
\newcommand{\Finally}{\lozenge\,}
\newcommand{\Ever}{\Diamond\,}
\newcommand{\Next}{\bigcirc\,}
\newcommand{\Until}{\mbox{$\, {\sf U}\,$}}

\newcommand{\mdp}{M}
\newcommand{\MdpInit}[1][]{\ensuremath{\mdp{#1}=\allowbreak(S{#1},\Act,\probmdp{#1})}}

\newcommand{\probmdp}{\mathcal{P}}

\newcommand{\fsc}{\ensuremath{\mathcal{A}}}

\newcommand{\ObsSym}{{Z}}
\newcommand{\ObsFun}{{O}}
\newcommand{\obs}{\ensuremath{z}}
\newcommand{\PomdpInit}[1][]{\pomdp{#1}=(\mdp{#1},\ObsSym{#1},\ObsFun{#1})}
\newcommand{\pomdp}{\mathcal{M}}

\newcommand{\states}{\ensuremath{S}}

\renewcommand{\Pr}{\ensuremath{\textnormal{Pr}}}

\newcommand{\sched}{\ensuremath{\gamma}}

\newcommand{\osched}{\ensuremath{\mathit{\gamma}}}
\newcommand{\oSched}{\ensuremath{\Gamma}}

\newcommand{\Act}{\ensuremath{\mathit{Act}}}
\newcommand{\act}{\ensuremath{a}}

\newcommand{\obsSeq}{\mathsf{ObsSeq}_{\mathit{fin}}}
\newcommand{\obsSeqFin}{\obsSeq}
\newcommand{\pathset}{\mathsf{Paths}}

\newcommand{\pathsfin}{\pathset_{\mathit{fin}}}

\newcommand{\last}[1]{\mathrm{last}(#1)}

\DeclareMathAlphabet{\mathpzc}{OT1}{pzc}{m}{it}
\def\presuper#1#2%
  {\mathop{}%
   \mathopen{\vphantom{#2}}^{#1}%
   \kern-\scriptspace%
   #2}

\newcommand{\RNNfun}{\ensuremath{\hat{\sched}}}

\begin{document}
\title{Verifiable RNN-Based Policies for POMDPs Under Temporal Logic Constraints}

\author{Steven Carr\textsuperscript{1}\thanks{Contact Authors}, Nils Jansen\textsuperscript{2}\footnotemark[1],Ufuk Topcu\textsuperscript{1}\\
  \affiliations \textsuperscript{1}The University of Texas at Austin \\
  \affiliations \textsuperscript{2}Radboud University, Nijmegen, The Netherlands
\emails stevencarr@utexas.edu, n.jansen@science.ru.nl\\
}

\maketitle
\acrodef{LSTM}[LSTM]{long short-term memory}
\acrodef{POMDP}[POMDP]{partially observable Markov decision process}
\acrodefplural{POMDP}[POMDPs]{partially observable Markov decision processes}
\acrodef{RNN}[RNN]{recurrent neural network}
\acrodef{FSC}[FSC]{finite-state controller}
\acrodef{DTMC}[MC]{discrete-time Markov chain}
\acrodef{LP}[LP]{linear programming}
\acrodef{FSC}[FSC]{finite-state controller}
\acrodef{MDP}[MDP]{Markov decision process}
\acrodefplural{MDP}[MDPs]{Markov decision processes}
\acrodef{LTL}[LTL]{linear-time temporal logic}
\acrodef{NN}[NN]{neural network}
\acrodef{QBN}[QBN]{quantized bottleneck network}
\acrodef{SMT}[SMT]{satisfiability-modulo-theories}
\acrodef{MILP}[MILP]{mixed-integer linear program}
\acrodef{ReLU}[ReLU]{rectified linear unit}
\acrodefplural{ReLU}[ReLUs]{rectified linear units}
\begin{abstract}
\acused{RNN}
Recurrent neural networks (RNNs) have emerged as an effective representation of control policies in sequential decision-making problems.
However, a major drawback in the application of \ac{RNN}-based policies is the difficulty in providing formal guarantees on the satisfaction of behavioral specifications, e.g. safety and/or reachability.
By integrating techniques from formal methods and machine learning, we propose an approach to automatically extract a \ac{FSC} from an \ac{RNN}, which, when composed with a finite-state system model, is amenable to existing formal verification tools.
Specifically, we introduce an iterative modification to the so-called quantized bottleneck insertion technique to create an \ac{FSC} as a randomized policy with memory.
For the cases in which the resulting \ac{FSC} fails to satisfy the specification, verification generates diagnostic information.
We utilize this information to either adjust the amount of memory in the extracted \ac{FSC} or perform focused retraining of the \ac{RNN}.
While generally applicable, we detail the resulting iterative procedure in the context of policy synthesis for \acp{POMDP}, which is known to be notoriously hard.
The numerical experiments show that the proposed approach outperforms traditional POMDP synthesis methods by 3 orders of magnitude within 2\% of optimal benchmark values.
\end{abstract}

\acresetall
\section{Introduction}
Research in the reinforcement and supervised learning communities has demonstrated the utility of \acp{RNN} in synthesizing control policies in domains that exhibit temporal behavior~\cite{DBLP:journals/ijon/TsoiB97,bakker2002reinforcement,DBLP:journals/corr/HeessHLS15}.
The internal memory states of \acp{RNN}, such as in \ac{LSTM} architectures~\cite{hochreiter1997long}, effectively account for temporal behavior by capturing the history from sequential information~\cite{pascanu2013construct}.
Furthermore, in applications that suffer from incomplete information, \acp{RNN} leverage history to act as either a state or value estimator \cite{sorensen1997simplified,wierstra2007solving} or as a control policy \cite{hausknecht2015deep}.



In safety-critical systems such as autonomous vehicles, policies that are guaranteed to prevent unsafe behavior are necessary.
%
We seek to provide formal guarantees for policies represented by \acp{RNN} with respect to temporal logic~\cite{Pnueli77} or reward specifications.
Such a verification task is, in general, hard due to the complex, often non-linear, structures of \acp{RNN}~\cite{DBLP:journals/csl/MulderBM15}.
Existing work directly employs \ac{SMT}~\cite{DBLP:journals/corr/abs-1811-06029}
or \ac{MILP} solvers~\cite{DBLP:conf/aaai/AkintundeKLP19}, however, such methods not only scale exponentially in the number of variables but also rely on constructions using only \acp{ReLU}.

\begin{figure}[b!]
	\definecolor{bg}{HTML}{ddeedd}
\definecolor{comp}{HTML}{c2d4dd}
\definecolor{impl}{HTML}{b0aac0}
\definecolor{ligb}{HTML}{5E7FC6}
\definecolor{bodybl}{HTML}{85A1DC}
\definecolor{headbl}{HTML}{264C9C}
\definecolor{bgyel}{HTML}{FFDC6B}
\definecolor{bodyyel}{HTML}{FFE58F}
\definecolor{headyel}{HTML}{E9BB25}
\tikzstyle{decision} = [diamond, draw, fill=blue!20, 
text width=4.5em, text badly centered, node distance=3cm, inner sep=0pt]
\tikzstyle{block} = [rectangle, draw, fill=blue!20, 
text width=5em, text centered, rounded corners, minimum height=4em]
\tikzstyle{line} = [draw, -latex']
\tikzstyle{cloud} = [draw, ellipse,fill=red!20, node distance=3cm,
minimum height=2em]
\def\checkmark{\tikz\fill[scale=0.6](0,.35) -- (.25,0) -- (1,.7) -- (.25,.15) -- cycle;} 
\centering
\begin{tikzpicture}[every node/.style={text centered, shape=rectangle, rounded corners, text width=2.0cm, minimum height=1.0cm, inner sep=2pt}]
\node[] (rnn) at (0,0) {RNN-based \\policy};

\node[below=1.0cm of rnn] (extract) {Extraction as an FSC};
\node[below=1.0cm of extract] (verify) {Formal verification};
\node[below=0.6cm of verify,text width=2.5cm] (model){System model\\ (e.g., a POMDP)};
\node[left=3.5cm of extract,text width=1.5cm] (hold){Entropy test};
\coordinate[left=2.5cm of rnn] (improve);

\draw (rnn) edge[-latex', very thick] (extract);
\draw (extract) edge[-latex', very thick] (verify);
\draw (model) edge[-latex', very thick] (verify);
\path[line,-latex', very thick] (verify)-| node[draw=none,above right,text width=3.5cm]{Diagnostics on induced behavior} (hold);
\path[line,-latex', very thick] (hold) -- node[below]{Increase precision} (extract);
\path[line,-latex',dashed,very thick] (hold) |- node[below right,text width=3.5cm]{Critical information for network retraining} (rnn);
\end{tikzpicture}
	\caption{High-level iterative policy extraction process.}
	\label{fig:high-level}
\end{figure}
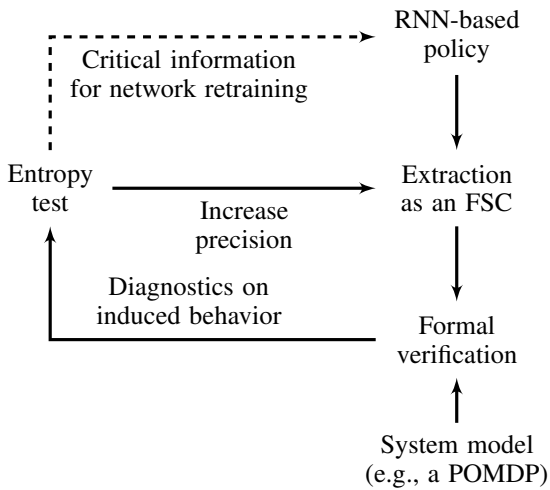

We take an iterative and model-based approach, summarized in Fig.~\ref{fig:high-level}.
In particular, we extract a policy in the form of a so-called \ac{FSC}~\cite{poupart2004bounded} from a given \ac{RNN}.
First, we employ a modification of a discretization technique called \emph{quantized bottleneck insertion}, introduced in~\cite{koul2018learning}.
Basically, the discretization facilitates a mapping of the continuous memory structure of the \ac{RNN} to a pre-defined number of discrete memory states and transitions of an \ac{FSC}.

However, this standalone \ac{FSC} is not sufficient to prove meaningful properties. 
The proposed approach relies on the exact behavior a policy induces on a specific application, e.g. construction of verifiable policies for \acp{POMDP}.
We apply the extracted \ac{FSC} directly to a formal model, e.g. a \ac{POMDP} that represents such an application. 
The resulting restricted model is amenable for efficient verification techniques that check whether a specification is satisfied~\cite{BK08}.

If the specification does not hold, verification methods typically provide diagnostic information on critical parts of the model in the form of so-called counterexamples.
%
%
%
%
%
We propose to utilize such counterexamples to identify \emph{improvements} in the extracted \ac{FSC} or in the underlying RNN.
First, increasing the amount of memory states in the \ac{FSC} may help to approximate the behavior of the \ac{RNN} more precisely~\cite{koul2018learning}.
Second, the \ac{RNN} may actually require further training data to induce higher-quality policies for the particular application.
Existing approaches rely, for example, on loss visualization~\cite{goodfellow2014qualitatively}, but we strive to exploit the information we can gain from the concrete behavior of the \acp{RNN} with respect to a formal model.
Therefore, in order to decide whether more data are needed in the training of the \ac{RNN} or whether first the number of memory states in the \ac{FSC} should be increased, we identify those critical decisions of the current \ac{FSC} that are ``arbitrary''. 
Basically, we measure the entropy~\cite{cover2012elements,biondi2013maximizing} of each stochastic choice over actions the current \ac{FSC}-based policy makes at critical states.
That is, if the entropy is high, the decision is deemed arbitrary despite its criticality and further training is required.



\acused{POMDP}
We showcase the applicability of the proposed method on \acp{POMDP}. 
With their ability to represent sequential decision-making problems under uncertainty and incomplete information, these models are of particular interest in planning and control~\cite{cassandra1998survey}.
Despite their utility as a modeling formalism and recent algorithmic advances, policy synthesis for \acp{POMDP} is hard both theoretically and practically~\cite{meuleau1999solving}.
For reasons outlined earlier, \acp{RNN} have recently emerged as efficient policy representations for \acp{POMDP}~\cite{hausknecht2015deep}, but the task of verifying the induced behavior is \emph{far more difficult than that for \acp{FSC}}.
We detail the proposed approach on \acp{POMDP} and combine the scalability and flexibility of an \ac{RNN} representation with the rigor of formal verification to synthesize \ac{POMDP} policies that adhere to temporal logic specifications.

We demonstrate the effectiveness of the proposed synthesis approach on a set of \ac{POMDP} benchmarks.
These benchmarks allows for a comparison to well-known POMDP solvers, both with and without temporal logic specifications.
The numerical examples show that the proposed method (1) is more scalable, by up to 3 orders of magnitude, than well-known POMDP solvers and (2) achieves higher-quality results in terms of the measure of interest than other synthesis methods that extract \acp{FSC}.

%

\paragraph{Related work.}
Closest to the proposed method is~\cite{carr2019counterexample}, which introduced a verification-guided method to train \acp{RNN} as POMDP policies. 
In particular, \cite{carr2019counterexample} extracts polices from the \acp{RNN} but unlike the proposed method, the extracted policies do not directly exhibit the memory structure of the \acp{RNN} and were instead handcrafted based on knowledge about the particular application.

%


There are three lines of related research. 
The first one concerns the formal verification of neural network-based control policies. 
Two prominent approaches~\cite{DBLP:conf/cav/HuangKWW17,katz2017reluplex} for the class of feed-forward deep neural networks  rely on encoding neural networks as \ac{SMT} problems through adversarial examples or \acp{ReLU} architectures respectively. 
\cite{DBLP:conf/aaai/AkintundeKLP19} concerns the direct verification of \acp{RNN} with \ac{ReLU} activation functions using SMT or MILP.
However, the scalability of these solver-based methods suffer from the size of the input models.
We circumvent this shortcoming by our model-based approach where verification is restricted to concrete applications followed by potential improvement of the \acp{RNN}.

The second relevant direction concerns the direct synthesis of \acp{FSC} for \acp{POMDP} without neural networks. 
For example, \cite{meuleau1999solving} uses a branch-and-bound method to compute optimal \acp{FSC} for POMDPs.
\cite{DBLP:conf/aaai/ChatterjeeCD16} uses a SAT-based approach to compute \acp{FSC} for qualitative properties.
\cite{junges2018finite} constructs an \ac{FSC} using parameter synthesis for Markov chains.

Third, existing work that concerns the extraction of \acp{FSC} from neural networks~\cite{Zeng:1993:LFM:188045.188064,pmlr-v80-weiss18a,finnegan2017maximum,michalenko2019representing},  does not integrate with formal verification to provide guarantees for extracted policies or to generate diagnostic information.
%
%

\section{Preliminaries}
\label{sec:prelim}
\noindent A \emph{probability distribution} over a set $\distDom$
is a function $\distFunc\colon\distDom\rightarrow\Ireal$ with $\sum_{\distDomElem\in\distDom}\distFunc(\distDomElem)=\distFunc(\distDom)=1$.
The set of all distributions on $\distDom$ is $\Distr(\distDom)$. The support of a distribution $\distFunc$ is
$\supp(\distFunc) = \{x\in\distDom\,|\,\distFunc(x)>0\}$.
The entropy of a distribution  $\distFunc$ is $\mathcal{H}(\mu) \colonequals - \sum_{\distDomElem \in \distDom} \mu(\distDomElem) \log_{|\distDom|} \mu(\distDomElem)$.
\paragraph{POMDPs.}
  A \emph{\ac{MDP}} $\mdp$ is a tuple $\MdpInit$ with
  a finite (or countably infinite) set $\states$ of \emph{states},
  a finite set $\Act$ of \emph{actions},
  and a \emph{transition probability function} $\probmdp\colon \states\times\Act\rightarrow\Distr(\states)$.
The reward function for states and actions is given by $\rew\colon\states\times\Act\rightarrow\R$.
A finite \emph{path} $\pi$ of an \ac{MDP} $\mdp$ is a sequence of states and actions; $\last{\pi}$ is the last state of $\pi$. 
The set of all finite paths is $\pathsfin^{\mdp}$.
%
\begin{definition}[POMDP]
	\label{def:pomdp}
	A \emph{\ac{POMDP}} is a tuple $\PomdpInit$, with $\mdp$ the \emph{underlying MDP of $\pomdp$}, $\ObsSym$ a finite set of observations, and $\ObsFun\colon\states\rightarrow\ObsSym$ the \emph{observation function}.
\end{definition}
\noindent For POMDPs, observation-action sequences are based on a finite path $\pi\in\pathsfin^{\mdp}$ of $\mdp$ and have the form: 
$\ObsFun(\pi)=\ObsFun(s_0)\xrightarrow{\act_0} \ObsFun(s_1)\xrightarrow{\act_1}\cdots\ObsFun(s_n)$.
The set of all finite observation-action sequences for a \ac{POMDP} $\pomdp$ is $\obsSeqFin^\pomdp$. 
\begin{definition}[POMDP Policy]
	\label{def:obsstrategy}
	An \emph{observation-based policy} for a \ac{POMDP} $\pomdp$ is a function $\osched\colon\obsSeqFin^\pomdp\rightarrow\Distr(\Act)$
	such that $\supp\bigl(\osched(\ObsFun(\pi))\bigr) \subseteq \Act\bigl(\last{\pi}\bigr)$ for all $\pi\in \pathsfin^{\mdp}$.
	$\oSched^\pomdp_\obs$ is the set of observation-based policies for $\pomdp$.
\end{definition}
A policy for a \ac{POMDP} resolves the nondeterministic choices in the \ac{POMDP}, based on the history of previous observations, by assigning distributions over actions. 
A \emph{memoryless} observation-based policy $\osched\in\oSched^\pomdp_\obs$ is given by $\osched\colon Z\rightarrow\Distr(\Act)$, \ie decisions are based on the current observation only.
A
 \ac{POMDP} $\pomdp$ together with a policy $\osched$ yields an \emph{induced \ac{DTMC}} $\pomdp^\osched$. 
 An MC does not contain any nondeterminism or partial observability.
 
Our definition restricts \ac{POMDP} policies to finite memory, which are typically represented as \acp{FSC}.

\begin{definition} [Finite-state controller (\ac{FSC})]
	\label{def:fsc}
	A $k$-\ac{FSC} for a \ac{POMDP} is a tuple $\mathcal{A} = (N,n_I,\alpha, \delta)$ where $N$ is a \emph{finite set} of $k$ memory nodes, $n_I \in N$ is the initial memory node, $\alpha$ is the action mapping $\alpha\colon N \times \ObsSym \rightarrow \Distr(\Act)$ and $\delta$ is the memory update $\delta\colon N\times \ObsSym \times \Act \rightarrow N	$.
	\end{definition}
An FSC has the observations $\ObsSym$ as input and the actions $\Act$ as output.
Upon an observation, depending on the current memory node the FSC is in, the action mapping $\alpha$ returns a distribution over $\Act$ followed by a change of memory nodes according to $\delta$.
\acp{FSC} are an extension of so-called \emph{Moore machines}~\cite{moore1956gedanken}, where the action mapping is deterministic, that is, $\alpha\colon N \times \ObsSym \rightarrow \Act$, and the memory update $\delta\colon N\times \ObsSym \rightarrow N$ does not depend on the choice of action.

\begin{definition}[Specifications]
\label{ssec:spec}
We consider \ac{LTL} properties~\cite{Pnueli77}. 
For a set of atomic propositions $AP$, which are either satisfied or violated by a state, and $a\in AP$, the set of LTL formulas is:
\[
\Psi\coloncolonequals a \ |\  (\Psi\land\Psi)\ |\ \neg\Psi\ |\ \Next\Psi\ |\ \Always\Psi\ |\ (\Psi\Until\Psi)\,.
\]
\end{definition}
Intuitively, a path $\pi$ satisfies the proposition $a$ if its first state does; $(\ltlformula_1\land\ltlformula_2)$ is satisfied,
if $\pi$ satisfies both $\ltlformula_1$ and $\ltlformula_2$; $\neg\ltlformula$ is true on $\pi$ if $\ltlformula$ is not satisfied.
The formula $\Next\ltlformula$ holds on $\pi$ if the subpath starting at the second state of $\pi$ satisfies $\ltlformula$; $\pi$ satisfies $\Always\ltlformula$ if all suffixes of $\pi$ satisfy $\ltlformula$. Finally, $\pi$ satisfies
$(\ltlformula_1\Until\ltlformula_2)$ if there is a suffix of $\pi$ that satisfies $\ltlformula_2$ and all longer suffixes satisfy
$\ltlformula_1$. $\Finally\ltlformula$ abbreviates $(\mathrm{true}\Until\ltlformula)$.

For POMDPs, one wants to synthesize a policy such that the probability of satisfying an LTL-property respects a given bound, denoted $\reachPropSymbol = \p_{\sim \lambda}(\ltlformula)$ for ${\sim}\in\{ {<}, {\leq},{\geq},{>}\}$ and $\lambda\in[0,1]$.
In addition, \emph{undiscounted expected reward properties} 
$\reachPropSymbol=\Ex_{\sim \lambda}(\Finally a)$ require that the expected accumulated cost until reaching a state satisfying $a$ respects $\lambda\in\R_{\geq 0}$.

A \emph{specification} $\reachPropSymbol$ is \emph{satisfied} for POMDP $\pomdp$ and $\sched$ if it is satisfied in the \ac{DTMC} $\pomdp^\osched$ ($\pomdp^\sched\models\reachPropSymbol$).


\paragraph{Policy network.} 
We now define a general notion of an \ac{RNN} that represents a POMDP policy.
\begin{definition}[Policy network]\label{def:policy_network}
	A policy network for a \ac{POMDP} is a function $\RNNfun\colon \obsSeqFin^\pomdp \rightarrow \Distr(\Act)$. 
\end{definition}
The underlying \ac{RNN} which receives sequential input in the form of (finite) observation sequences from $\obsSeqFin^\pomdp$, the output is a distribution over actions, see Fig.~\ref{fig:RNN_policy}.
To be more precise, we identify the main components of such a network.
\begin{definition}[Components of a policy network]\label{def:policy_network_components}
	A policy network $\RNNfun$ is sufficiently described by a \emph{hidden-state update function} 
	$\hat{\delta} \colon \Reals \times \ObsSym \times \Act \rightarrow \Reals$ and an \emph{action mapping} $\gamma_h \colon \Reals \rightarrow \Distr(\Act)$.
\end{definition}
Consider the following observation sequence:
\begin{align}
\ObsFun(\pi)=\ObsFun(s_0)\xrightarrow{\act_0} \ObsFun(s_1) \xrightarrow{\act_1}\cdots \ObsFun(s_i)
\label{eq:obs_seq}
\end{align}
The policy network receives an observation and returns an action choice.
Throughout the execution of the sequence, the \ac{RNN} holds a continuous hidden state $h\in \Reals$, occasionally described as an internal memory state, which captures previous information.
On each transition, this hidden state is updated to include the information of the current state and the last action taken under the hidden state transition function $\hat{\delta}$.
From the prior observation sequence in (\ref{eq:obs_seq}), the corresponding hidden state sequence would be defined as:
\begin{align*}
\hat{\delta}(\pi) = h_0 \xrightarrow{\act_0,~\ObsFun(s_1)} h_1  \xrightarrow{\act_1,~\ObsFun(s_2)} \cdots h_i
\end{align*}
Additionally, the output of the policy network is expressed by the action-distribution function $\gamma_h$, which maps the value of hidden state to a distribution over the actions.
At internal memory states $h_i$, we have $\hat{\delta}(h_i,\ObsFun(s_i),a_i) = h_{i+1}$ and $\gamma_h(h_{i+1})=\mu(\Act)$ for state $s_i$ on path $\pi$.
Note that a policy network characterizes a well-defined POMDP policy.

\section{Problem Statement}
We attempt to solve two separate but related problems:
(1) For a POMDP $\pomdp$, a policy network $\RNNfun$  and a specification $\varphi$, 
the problem is to extract an \ac{FSC} $\fsc_{\hat{\gamma}} \in \oSched^\pomdp_\obs$ such that  $\pomdp^{\fsc_{\hat{\gamma}}} \models \varphi$. 
(2) If the extraction process fails to produce a suitable candidate, 
then we improve the policy network $\RNNfun^*$ for which we can solve (1).

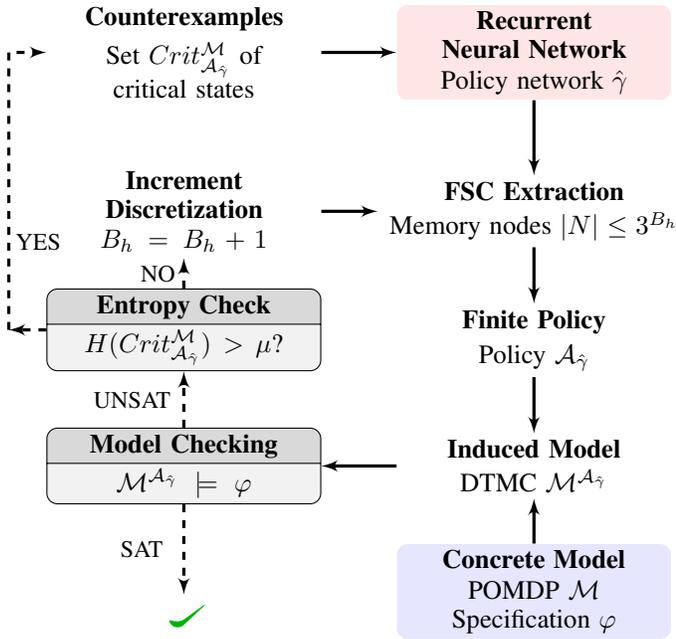
\begin{figure}
	\definecolor{bg}{HTML}{ddeedd}
\definecolor{comp}{HTML}{c2d4dd}
\definecolor{impl}{HTML}{b0aac0}
\definecolor{ligb}{HTML}{5E7FC6}
\definecolor{bodybl}{HTML}{85A1DC}
\definecolor{headbl}{HTML}{264C9C}
\definecolor{bgyel}{HTML}{FFDC6B}
\definecolor{bodyyel}{HTML}{FFE58F}
\definecolor{headyel}{HTML}{E9BB25}
\tikzstyle{decision} = [diamond, draw, fill=blue!20, 
text width=4.5em, text badly centered, node distance=3cm, inner sep=0pt]
\tikzstyle{block} = [rectangle, draw, fill=blue!20, 
text width=5em, text centered, rounded corners, minimum height=4em]
\tikzstyle{line} = [draw, -latex']
\tikzstyle{cloud} = [draw, ellipse,fill=red!20, node distance=3cm,
minimum height=2em]
\def\checkmark{\tikz\fill[scale=0.6](0,.35) -- (.25,0) -- (1,.7) -- (.25,.15) -- cycle;} 
\centering
\begin{tikzpicture}[every node/.style={text centered, shape=rectangle, rounded corners, text width=4cm, minimum height=1.0cm, inner sep=2pt}]
\tikzset{
	normalnode/.style={text width=3.5cm,
		fill=red!10
	}
}
\tikzset{
	splitnode/.style={
		rectangle split,
		rectangle split parts=2,
		text width=3.5 cm
	}}
\tikzset{
		splitnodefill/.style={draw,
			rectangle split,
			rectangle split parts=2,
			rectangle split part fill={gray!30,gray!10},
			text width=3.5cm
		}}
\tikzstyle{split}=[rectangle split,rectangle split parts=2]

\node[splitnode](train){};
\node[splitnode, fill=red!10] (rnn) at (0,0) {\textbf{Recurrent\\Neural Network} \nodepart{second} Policy network $\RNNfun$};

\node[splitnode,below=1.0cm of rnn,text width=4.0 cm](fsc) {\textbf{FSC Extraction} \nodepart{second} Memory nodes $|N|\leq 3^{B_h}$};
\node[splitnode,below=0.75cm of fsc] (strategy) {\textbf{Finite Policy} \nodepart{second} Policy $\fsc_{\hat{\osched}}$};
\node[splitnode, below=0.75cm of strategy] (dtmc) {\textbf{Induced Model} \nodepart{second} DTMC $\pomdp^{\fsc_{\hat{\gamma}}}$ };
\node[splitnodefill, left=1.0cm of dtmc] (model) {\textbf{Model Checking} \nodepart{second} $\pomdp^{\fsc_{\hat{\gamma}}} \models \varphi$ };
\node[splitnodefill,above=0.75cm of model](entropy){\textbf{Entropy Check} \nodepart{second} $H(Crit^\pomdp_{\fsc_{\hat{\gamma}}})>\mu$?};
\node[splitnode,left=0.75cm of fsc](count) {\textbf{Increment \\ Discretization} \nodepart{second} $B_h = B_h + 1$};

\node[splitnode,left=1.0cm of train](counter) {\textbf{Counterexamples} \nodepart{second} Set $Crit^\pomdp_{\fsc_{\hat{\gamma}}}$ of critical states};





\node[splitnode, below=0.6cm of dtmc,fill=blue!10] (pomdp) {\textbf{Concrete Model} \nodepart{second}POMDP $\pomdp$ \\ Specification $\varphi$};

\node[draw=none,below=1.2cm of model,minimum height=0.1cm,text width =0.5cm] (safe) {\color{green!70!black}\Large\checkmark};


\path [line,-latex', very thick, dashed] (model) --node[draw=none,left,text width=1.2cm]{\small UNSAT } (entropy);
\path [line,-latex', very thick, dashed] (entropy) --node[draw=none,left,text width=0.5cm]{\small NO } (count);
\path [line,-latex', very thick, dashed] (model) --node[draw=none,left=0.5cm,text width=0.25cm,minimum height=0.05cm]{\small SAT} (safe);
\coordinate[left=0.5cm of entropy.west] (c1);
\path [line,-latex', very thick, dashed] (entropy.west) -- (c1);
\path [line,-latex', very thick, dashed] (c1) |- node[draw=none,below right=2cm and 0cm,text width=0.5cm]{\small YES } (counter);
\draw (rnn) edge[-latex', very thick] (fsc);
\draw (fsc) edge[-latex', very thick] (strategy);
\draw (counter) edge[-latex', very thick] (rnn);
\draw (count) edge[-latex',very thick] (fsc);
\draw (strategy.south) edge[-latex', very thick] (dtmc.north);

\draw (dtmc.west) edge[-latex', very thick] (model.east);

\draw (pomdp) edge[-latex', very thick] (dtmc);
\end{tikzpicture}
	\caption{Procedural flow for the iterative FSC extraction and RNN-based policy improvement.}
	\label{fig:flowchart}
\end{figure}

\subsection{Outline}
Fig.~\ref{fig:flowchart} illustrates the workflow of the proposed approach for a given POMDP $\pomdp$, policy network $\RNNfun$ and specification $\varphi$. 
We summarize the individual steps below and provide the technical details in the subsequent sections.

\paragraph{\ac{FSC} extraction.}
We first quantize the memory nodes of the policy network $\hat{\gamma}$, that is, we discretize the memory update of the continuous memory state $h$.
From this discrete representation of the memory update, we construct an \ac{FSC} $\fsc_{\hat{\gamma}} \in \oSched^\pomdp_\obs$.
The procedure has as input the number $B_h$ of neurons which defines a bound on the number of memory nodes in the FSC.

\paragraph{Verification.}
We use the \ac{FSC} $\fsc_{\hat{\gamma}}$  to resolve partial information and nondeterministic choices in the \ac{POMDP} $\pomdp$, resulting in an induced \ac{DTMC} $\pomdp^{\fsc_{\hat{\gamma}}}$.
We evaluate whether the given specification $\varphi$ is satisfied for this induced \ac{DTMC} using a formal verification technique called \emph{model checking}~\cite{BK08}.
If the specification $\varphi$ holds, then the synthesis is complete with output policy $\fsc_{\hat{\gamma}}$.
However, if $\varphi$ does not hold, then we decide if we shall increase the bound $B_h$ on the number of memory nodes or if the network needs retraining. 
In particular, we examine whether or not the entropy over the \ac{FSC}'s action distribution is above a prescribed threshold.

\paragraph{Policy improvement.}
In the high entropy case, we increase the discretization level, that is, we increase $B_h$, and construct the \ac{FSC} $\fsc_{\hat{\gamma}}$ with additional memory states at its disposal.
Whereas in the other case, additional memory nodes may cause the extracted \ac{FSC} to be drawn from extrapolated information and we instead seek to improve the policy network. 
For that, we use diagnostic information in the form of counterexamples to generate new data~\cite{carr2019counterexample}.

\begin{figure}
	\subfloat[5-State POMDP\label{fig:pomdp}]{
	\centering
\begin{tikzpicture}[scale=1.5, state/.append style={minimum size=2mm,inner sep=3pt},>=stealth,
        bobbel/.style={minimum size=1.5mm,inner sep=0pt,fill=black,circle}]
    \node[state,fill=blue!15!white] (s0) at (0.5,0) {$s_0$};
    \node[bobbel] (a0) at (-0.5,0) {};
    \draw[<-] (a0) -- +(-0.5,0);
    \draw (a0) edge[->] node[above,pos=0.75]{$1/3$}(s0);
    \node[state,fill=blue!15!white] (s1) at (1.5,0.5) {$s_1$};
    \draw (a0) edge [->,bend left] node[above,pos=0.35]{$1/3$} (s1);
    \node[state,fill=blue!15!white] (s2) at (1.5,-0.5) {$s_2$};
     \draw (a0) edge [->,bend right] node[below,pos=0.35]{$1/3$} (s2);
    \node[state] (s4) at (3.0,-0.5) {$s_4$};
    \node[state,accepting] (s3) at (3.0,0.5) {$s_3$};

    \draw (s0) edge node[below,pos=0.65]{up} (s1);
    \draw (s0) edge node[above,pos=0.65]{down} (s2);
    \draw (s2) edge[->] node[below,pos=0.65]{down} (s4);
    \draw (s1) edge[->,loop above] node[right,pos=0.75]{up} (s1);
    \draw (s2) edge[->] node[above left,pos=0.5]{up} (s3);
    \draw (s1) edge[->] node[above right,pos=0.25]{down} (s3);
    \draw (s3) edge[->,loop right] node[right,pos=0.5]{\ \act} (s3);
    \draw (s4) edge[->,loop right] node[right,pos=0.5]{\ \act} (s4);
  \end{tikzpicture}
} \newline
\centering
\subfloat[1-FSC \label{fig:1FSC}]{
\begin{tikzpicture}[scale=1.5, state/.append style={minimum size=2mm,inner sep=3pt},>=stealth,
bobbel/.style={minimum size=2.5mm,inner sep=0pt,fill=black,circle}]
\node[state,fill=blue!15!white] (f0) at (0,0) {$(0,blue)$};
\node[bobbel,label=right:{up}] (a0) at (0,1.05) {};
\node[bobbel,label=right:{down}] (a1) at (0,-1.05) {};
\coordinate[] (xin) at (-1,0);
\draw (f0.north) edge[->,bend left] node[left]{$p$} (a0);
\draw (f0.south) edge[->,bend right] node[left]{$1-p$} (a1);
\draw (a0) edge[->,bend left] node[left]{} (f0.north);
\draw (a1) edge[->,bend right] node[left]{} (f0.south);
\draw (xin) edge[->] (f0.west);

\end{tikzpicture}}\hfill
\subfloat[2-FSC \label{fig:2FSC}]{
\begin{tikzpicture}[scale=1.5, state/.append style={minimum size=2mm,inner sep=3pt},>=stealth,
bobbel/.style={minimum size=2.5mm,inner sep=0pt,fill=black,circle}]
\node[state,fill=blue!15!white] (f0) at (0,0) {$(0,blue)$};
\node[state,fill=blue!15!white] (f1) at (2,0) {$(1,blue)$};
\coordinate[] (xin) at (-1,0);

\node[bobbel,label=below:{up}] (a0) at (1,1.05) {};
\node[bobbel,label=above:{down}] (a1) at (1,-1.05) {};

\draw (f0.north) edge[->,bend left] node[above left]{$1$} (a0);
\draw (f1.south) edge[->,bend left] node[below right]{$1$} (a1);
\draw (a0) edge[->,bend left] node[left]{} (f1.north);
\draw (a1) edge[->,bend left] node[left]{} (f0.south);
\draw (xin) edge[->] (f0.west);

\end{tikzpicture}
}
	\caption{(a) POMDP for Example~\ref{ex:motivating} with (b) $1$-FSC and (c) $2$-FSC. Both FSCs are defined for observing ``blue'' and subsequent action choices that may result in a change of memory node for the $2$-FSC.}
	\label{fig:motivating}
\end{figure}
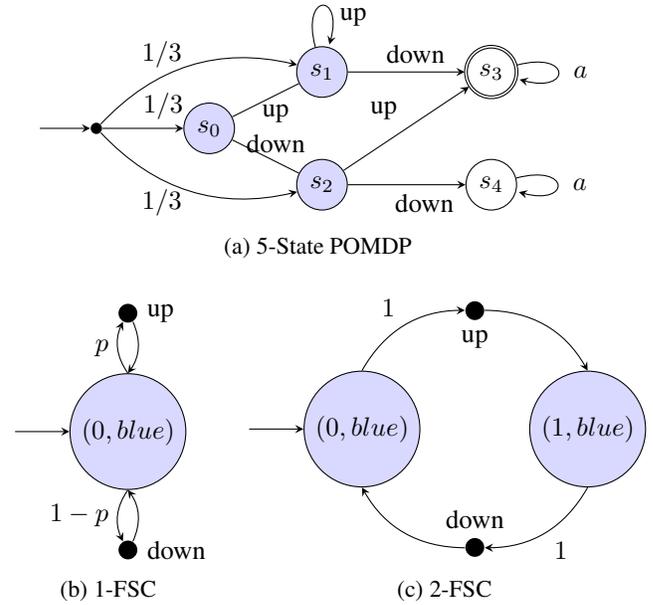

\begin{example}
	\label{ex:motivating}
	We consider the POMDP in Fig.~\ref{fig:motivating} as a motivating example for the necessity of memory-based \acp{FSC}.
	The \ac{POMDP} has three observations (``blue'', $s_3$ and $s_4$) where observation ``blue'' is received upon visiting $s_0$, $s_1$, and $s_2$. 
	That is, the agent is unable to distinguish between these states.
	The specification is $\varphi = \Pr_{\geq 0.9}(\Ever s_3)$, so the agent is to reach state $s_3$ with at least probability $0.9$.
	In a $1$-\ac{FSC} (i.e. one memory node $0$), we can describe an \ac{FSC} $\fsc_1$ by:
	\begin{align*}
	\alpha(0,blue) &= \begin{cases} up & \textrm{with probability} \quad p,	\\ down & \textrm{with probability} \quad 1-p,\end{cases}\\
	\delta(0,\obs,\act) &= 0 \quad \forall \obs \in \ObsSym, \act \in \Act.
	\end{align*}
	A $2$-\ac{FSC} with two memory nodes ($0$ and $1$), see Fig.~\ref{fig:2FSC}, allows for greater expressivity, i.e. the policy can base its decision on larger observation sequences. With this memory structure, we can create an \ac{FSC} $\fsc_2$ that ensures the satisfaction of $\varphi$:
	\begin{align*}
		\alpha(0,blue) &= \begin{cases} up & \textrm{with probability} \quad 1,	\\ down & \textrm{with probability} \quad 0,\end{cases}\\
		\alpha(1,blue) &= \begin{cases} up & \textrm{with probability} \quad 0,	\\ down & \textrm{with probability} \quad 1,\end{cases}\\
		\delta(0,blue,up) &= 1, \\
		\delta(1,blue,down) &= 0. 
	\end{align*}
\end{example}

\section{Policy Extraction}
\label{ssec:extraction}
In this section we describe how we adapt the method called quantized bottleneck insertion~\cite{koul2018learning} to extract an \ac{FSC} from a given \ac{RNN}.
Let us first explain the relationship between the main components of a policy network $\RNNfun$ (Definition~\ref{def:policy_network_components}) and an \ac{FSC} $\fsc$ (Definition~\ref{def:fsc}).
In particular, the hidden-state update function $\hat{\delta} \colon \Reals \times \ObsSym \times \Act \rightarrow \Reals$ takes as input a real-valued hidden state of the policy network, while the memory update function of an \ac{FSC} takes a memory node from the finite set $N$.
The key for linking the two is therefore a mechanism that encodes the continuous hidden state $h$ into a set $N$ of discrete memory nodes.

\begin{figure}[t]
	\subfloat[RNN Policy. \label{fig:RNN_policy}]{\hspace{-1.0cm}\scalebox{0.8}{\centering\definecolor{bg}{HTML}{ddeedd}
\definecolor{comp}{HTML}{c2d4dd}
\definecolor{impl}{HTML}{b0aac0}
\definecolor{ligb}{HTML}{5E7FC6}
\definecolor{bodybl}{HTML}{85A1DC}
\definecolor{headbl}{HTML}{264C9C}
\definecolor{bgyel}{HTML}{FFDC6B}
\definecolor{bodyyel}{HTML}{FFE58F}
\definecolor{headyel}{HTML}{E9BB25}
\tikzstyle{decision} = [diamond, draw, fill=blue!20, 
text width=4.5em, text badly centered, node distance=3cm, inner sep=0pt]
\tikzstyle{block} = [rectangle, draw, fill=blue!20, 
text width=5em, text centered, rounded corners, minimum height=4em]
\tikzstyle{line} = [draw, -latex']
\tikzstyle{cloud} = [draw, ellipse,fill=red!20, node distance=3cm,
minimum height=2em]
\def\checkmark{\tikz\fill[scale=0.6](0,.35) -- (.25,0) -- (1,.7) -- (.25,.15) -- cycle;} 
\def\layersep{2.5cm}
\def\rnnlayersep{2.0cm}
\centering
\begin{tikzpicture}[every node/.style={draw, text centered, shape=rectangle, rounded corners, text width=1cm, minimum height=1.5cm, inner sep=5pt}]
\tikzset{
	splitnode/.style={
		rectangle split,
		rectangle split parts=2,
		rectangle split part fill={headbl!70,bodybl!60}
	}
}
\tikzset{
	normalnode/.style={
		fill=red!10
	}
}
\tikzstyle{annot} = [text width=2em, text centered, draw=none]
\tikzstyle{acts} = [circle,text width=0.05em,minimum height=0.15cm,inner sep=0pt, text centered, fill=black]
\tikzstyle{neuron}=[circle,fill=black!25,minimum size=6pt,inner sep=0pt]
\tikzstyle{rnn}=[rectangle,fill=black!25,minimum width=1.8cm,minimum height=1.2cm,inner sep=0pt]
\node[annot](o-in) at (0,-2.5){\LARGE $\obs$};
\node[rnn](LSTM) at (\layersep,-2.5){\LARGE RNN};
%
\path[->,very thick] (o-in) edge (LSTM);
\node[neuron](out-layer) at (2*\layersep,-2.5){\LARGE $\sigma$};
\path[->,very thick] (LSTM) edge (out-layer);
\path[->,very thick] (LSTM) edge [in=160,out=20,loop,looseness=4] (LSTM);
\foreach \name / \y in {1,...,3}
{\node[acts,pin={[pin edge={white,solid,<-},pin distance=0.05cm]right:\LARGE$a_\y$}](act-\name) at (3*\layersep,-0.5-\y){};
	\path[->,very thick] (out-layer) edge (act-\name);
}
\node[annot](inlay) at (0,0) {Input };
\node[annot,text width=2.0cm](lay)at(\layersep,0) {Recurrent Layer};
\node[annot,text width=2.0cm](lay)at(2*\layersep,0) {Softmax Layer};
\node[rectangle,fill=none,minimum width=2.4cm,minimum height=1.8cm,draw,dashed,very thick](box) at (\layersep,-2.5) {};
\end{tikzpicture}}}\\[-4ex]
	\subfloat[RNN block and associated \ac{QBN} of $B_h=3$ with quantized activation $\hat{\sigma} \colon \Reals\rightarrow \lbrace-1,0,1\rbrace$.	\label{fig:QBN}]{\hspace{-1.0cm}\scalebox{0.8}{\definecolor{bg}{HTML}{ddeedd}
\definecolor{comp}{HTML}{c2d4dd}
\definecolor{impl}{HTML}{b0aac0}
\definecolor{ligb}{HTML}{5E7FC6}
\definecolor{bodybl}{HTML}{85A1DC}
\definecolor{headbl}{HTML}{264C9C}
\definecolor{bgyel}{HTML}{FFDC6B}
\definecolor{bodyyel}{HTML}{FFE58F}
\definecolor{headyel}{HTML}{E9BB25}
\tikzstyle{decision} = [diamond, draw, fill=blue!20, 
text width=4.5em, text badly centered, node distance=3cm, inner sep=0pt]
\tikzstyle{block} = [rectangle, draw, fill=blue!20, 
text width=5em, text centered, rounded corners, minimum height=4em]
\tikzstyle{line} = [draw, -latex']
\tikzstyle{cloud} = [draw, ellipse,fill=red!20, node distance=3cm,
minimum height=2em]
\def\checkmark{\tikz\fill[scale=0.6](0,.35) -- (.25,0) -- (1,.7) -- (.25,.15) -- cycle;} 
\def\layersep{2.5cm}
\def\rnnlayersep{2.0cm}
\centering

\begin{tikzpicture}[every node/.style={draw, text centered, shape=rectangle, rounded corners, text width=1cm, minimum height=1.5cm, inner sep=5pt}]
\tikzstyle{annot} = [text width=2em, text centered, draw=none]
\tikzstyle{acts} = [circle,text width=0.05em,minimum height=0.15cm,inner sep=0pt, text centered, fill=black]
\tikzstyle{neuron}=[circle,fill=black!25,minimum size=6pt,inner sep=0pt]
\tikzstyle{rnn}=[rectangle,fill=black!25,minimum width=1.8cm,minimum height=1.2cm,inner sep=0pt]

\node[rnn](LSTM_mod) at (0.0*\rnnlayersep,-2.0){\LARGE RNN};
\node[annot] (h_out) at (0.65*\rnnlayersep,-1.75) {\LARGE $h$};
\node[circle,minimum size=0.2cm,minimum height=0.5cm,inner sep=0pt, text width=0.1cm] (en_in) at (1*\rnnlayersep,-2.0){};
\path[->,very thick] (LSTM_mod) edge (en_in);
\node[circle,minimum size=0.2cm,minimum height=0.5cm,inner sep=0pt, text width=0.1cm] (de_out) at (3.25*\rnnlayersep,-2.0){};
\foreach \name / \y in {1,...,3}
{\node[circle,minimum size=0.2cm,minimum height=0.5cm,inner sep=0pt, text width=0.1cm] (I-\name) at (1.75*\rnnlayersep,-0.0-\y) {};
	\path[->,very thick] (en_in) edge (I-\name);
	\node[circle,minimum size=0.2cm,minimum height=0.5cm,inner sep=0pt, text width=0.1cm] (O-\name) at (2.5*\rnnlayersep,-0.0-\y) {};
	\path[->,very thick] (I-\name) edge node[above,draw=none,text width=1.0cm,text height=1.25cm]{\LARGE $\hat{\sigma}$} (O-\name);
	\path[->,very thick] (O-\name) edge (de_out);
}
\node[annot](h-out) at (4.0*\rnnlayersep,-2.0){\LARGE $\hat{h}$};
\path[->,very thick] (de_out) edge (h-out);
\node[annot](inlay) at (1.0*\rnnlayersep,0) {Encoder};
\node[annot,text width=2.5cm](lay)at(2.75*\rnnlayersep,0) {Decoder};
\node[rectangle,fill=none,minimum width=2.4cm,minimum height=3.2cm,draw,dashed,very thick](e-box) at (2.28*\layersep,-2.0) {};
\node[rectangle,fill=none,minimum width=2.4cm,minimum height=3.2cm,draw,dashed,very thick](d-box) at (1.1*\layersep,-2.0) {};
\coordinate (x-under) at (4.0*\rnnlayersep,-4);
\path[->,very thick,draw] (h-out)  edge (x-under);
\path[->,very thick,draw] (x-under) -| (LSTM_mod);

\end{tikzpicture}}}
	\caption{Policy network structure without and with a QBN.}

\end{figure}
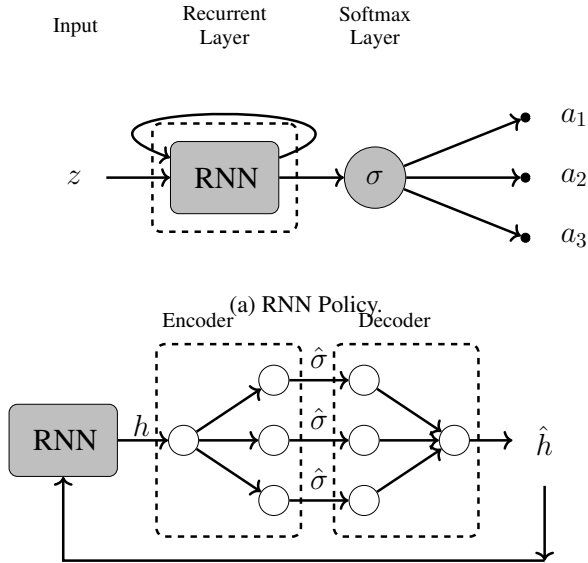

\paragraph{Policy network modification.}
To obtain the above linkage, we leverage an autoencoder~\cite{goodfellow2016deep} in the form of a \emph{quantized bottleneck network} (\ac{QBN})~\cite{koul2018learning}.
This QBN, consisting of an encoder and a decoder, is inserted into the policy network directly before the softmax layer, see Fig.~\ref{fig:QBN}. 
In the encoder, the continuous hidden state value $h\in\Reals$ is mapped to an intermediate real-valued vector $\Reals^{B_h}$ of pre-allocated size $B_h$.
The decoder then 
maps this intermediate vector into a discrete vector space defined by $\lbrace -1,0,1\rbrace^{B_h}$.
This process, illustrated in Fig.~\ref{fig:QBN}, provides a mapping of the continuous hidden state $h$ into $3^{B_h}$ possible discrete values.
We denote the discrete state for $h$ by $\hat{h}$ and the set of all such discrete states by $\hat{H}$. 
Note, that $|\hat{H}|\leq 3^{B_h}$ since not all values of the hidden state may be reached in an observation sequence.
\cite{koul2018learning} has another \ac{QBN} for a continuous observation space, however, we focus on discrete observations and can neglect the additional autoencoder.

\paragraph{FSC construction.}
After the \ac{QBN} insertion we simulate a series of executions, querying the modified \ac{RNN} for action choices, on the concrete application, e.g. using a \ac{POMDP} modelx.
We form a dataset of consecutive pairs ($\hat{h}_{t}, \hat{h}_{t+1})$ of discrete states, the action $\act_{t}$ and the observation $\obs_{t+1}$ that led to the transition $\lbrace \hat{h}_{t},\act_{t},\obs_{t+1},\hat{h}_{t+1}\rbrace$ at each time $t$ during the execution of the policy network.
The number of accessed memory nodes $N\subseteq\hat{H}$ corresponds to the number of different discrete states $\hat{h} \in \hat{H}$ in this dataset.  
The deterministic memory update rule $\delta(n_{t},\act_{t},\obs_{t+1}) = n_{t+1}$ is obtained by constructing a $N\times (|\ObsSym|\times |\Act|)$ transaction table, for a detailed description see \cite{koul2018learning}.
We can additionally construct the action mapping $\alpha\colon N \times \ObsSym \rightarrow \Distr(\Act)$ with $\alpha(n_t,\obs_t) = \mu\in\Distr(\Act)$ by querying the softmax-output layer (see Fig.~\ref{fig:RNN_policy}) for each memory state and observation.

\section{Policy Evaluation and Improvement}\label{sec:evaluation_improvement}
\setlength{\aboverulesep}{-1pt}
\setlength{\belowrulesep}{0pt}
\begin{table*}[t]
	\scriptsize\setlength{\tabcolsep}{3.5pt}
	\centering
	\scalebox{0.95}{%
		\begin{tabular}{@{}lrrc|rrr|rrr|rr|rr@{}}
			\toprule
			\multicolumn{4}{c|}{} & \multicolumn{3}{c|}{Extraction Approach}   & \multicolumn{3}{c|}{Handcrafted} & \multicolumn{2}{c|}{{PRISM-POMDP}} & \multicolumn{2}{c}{{SolvePOMDP}}\\
			Problem   & $|\states|$ & $|\ObsSym|$  & Type   & Memory & Value & Time (s)& Memory &  Value & Time (s) & Value & Time (s) & Value & Time (s) \\
			\midrule
			Maze(1) & 11 & 7  & Min  & 2 & 4.33 & 80.31 & 2 & 4.31 & 30.70 & \textbf{4.30} & \textbf{0.09} & 4.30 & 0.30 \\
			Maze(2) & 14& 7 & Min & 3 & 5.34 & 114.23 & 3  & 5.31& 46.65 & 5.23 & 2.176 & \textbf{5.23} & \textbf{0.67}\\
			Maze(5) & 23& 7 & Min & 3 &  13.29 & 160.12 & 6   & 14.40 & 68.09 & 13.00* & 4110.50 & \textbf{12.04} & \textbf{134.46} \\
			Maze(10) & 38& 7 & Min & 5 & \textbf{23.02} & 210.01 & 11 & 100.21 & \textbf{158.33}  & MO  & MO & MO & MO \\
			\rowcolor{gray!25} Grid(3) & 9 & 2 & Min & 3 & 2.90 & 87.31  &	2 & 2.90 & 38.94 & 2.88 & 2.332 & \textbf{2.88} & \textbf{0.06} \\
			\rowcolor{gray!25} Grid(4) & 16 & 2 & Min & 7 & 4.20 & 124.31 & 3 & 4.32 & 79.99 & 4.13	 & 1032.53 & \textbf{4.13} & \textbf{0.73}\\
			\rowcolor{gray!25} Grid(5) & 25 & 2 & Min & 9 & 5.91  & 250.14  &	4 & 6.623 & 91.42 & MO  & MO  &  \textbf{5.42} & \textbf{1.97}  \\
			\rowcolor{gray!25} Grid(10) & 100 & 2 & Min & 9 &\textbf{12.92} & 1031.21 & 9  & 13.63  & \textbf{268.40} & MO & MO & MO  & MO \\
			\rowcolor{gray!25} Grid(25) & 625 & 2 & Min & 16 &  \textbf{35.32} & 6514.30 & 24 &	531.05 &  \textbf{622.31} & MO  & MO & MO  & MO \\
			%
			%
			Navigation (4) & 256 & 256 & Max  & 8 & 0.92 & 160.32  & 8 & 0.92 & 80.26 & \textbf{0.93}*  & \textbf{1034.64} & NA & NA  \\
			 Navigation (5) & 625& 256 & Max  & 8 & \textbf{0.95} & 311.65   & 8 & 0.92 & \textbf{253.11} & MO  & MO & NA & NA \\
			 Navigation (10) & $10^4$& 256  & Max & 8 & \textbf{0.90} & 2561.02 & 4 & 0.85 & \textbf{1471.17} & MO & MO & NA & NA \\
			 Navigation (20) & $1.6\times 10^5$& 256  & Max & 9 & \textbf{0.98} & 8173.03 & 4 & 0.96 & \textbf{7068.24} & MO & MO & NA & NA \\
			\bottomrule
		\end{tabular}}
	\caption{Synthesizing strategies for examples with expected reward and LTL specifications.}
\label{tab:Res}
\end{table*}
\paragraph{Evaluation using formal verification.}
We assume that for \ac{POMDP} $\PomdpInit$ and specification $\varphi$, we have an extracted \ac{FSC} $\fsc_{\hat{\gamma}} \in  \oSched^\pomdp_\obs$ as in Definition~\ref{def:fsc}.
We use the policy $\fsc_{\hat{\gamma}}$ to obtain the induced \ac{DTMC} $\pomdp^{\fsc_{\hat{\gamma}}}$. 
For this \ac{DTMC}, formal verification through model checking checks whether $\pomdp^{\fsc_{\hat{\gamma}}}\models\varphi$ and thereby provides hard guarantees about the quality of the extracted \ac{FSC} $\fsc_{\hat{\gamma}}$ regarding $\varphi$.
Model checking provides the probability (or the expected reward) to satisfy a specification for \emph{all states} $s\in S$ via solving linear equation systems.

\addtocounter{example}{-1}
\begin{example}[cont.]
	Consider the case in the 1-FSC $\fsc_1$ (Fig.~\ref{fig:1FSC}) where $p=1$, the probability of reaching the state $s_3$ in the induced \ac{DTMC} is $\Pr(\Ever s_3) = \frac{1}{3}$.
	Clearly, the behavior induced by this 1-FSC violates the specification and formal verification provides two counterexamples of critical memory-state pairs for this policy $\fsc_{\hat{\gamma}}$: $(0,s_0)$ and $(0,s_1)$.
\end{example}
%
After model checking, if the specification does not hold, the policy may require refinement.
As discussed before, on the one hand we can increase the upper bound $B_h$ on the number of memory nodes to extract a new FSC. 
At each iteration of the inner loop in Fig.~\ref{fig:flowchart}, we modify the \ac{QBN} for the new, increased, level of discretization and obtain a new FSC using the process outlined in Sect~\ref{ssec:extraction}.
On the other hand, we may decide via a formal entropy check whether new data need to be generated to actually improve the policy.


\paragraph{Improving the policy network.}
\begin{figure}[t]
	\begin{tikzpicture}
		\begin{axis}[
		title = {Entropy of the Extracted FSC},
		xlabel = {Number of Samples},
		xmode=log,
		y label style={at={(axis description cs:0.1,.5)},rotate=0,anchor=south},
		ylabel = {Average Entropy $H(\osched_\fsc)$},
		ymin= 0,
		ymax=1,
		xmin=250,
		xmax=160000,
		label style={font=\small},
		tick label style={font=\small},
		width = 0.45\textwidth,
		height = 0.28\textwidth,
		legend style={anchor=north east,font=\small}]
		\addplot+[mark=*,only marks,mark options={scale = 1.0pt,color=green!20!black,fill=green!20!black}]coordinates{
			(408,0.86)(800,0.84)(1401,0.800)(2511,0.790)(4374,0.700)(8523,0.620)(12739,0.523)(25218,0.486)(37041,0.455)(49304,0.435)(61173,0.425)(76569,0.414)(92222,0.415)(107913,0.413)(150144,0.412)
		};\label{bh1}
		\addplot+[mark=square*,only marks,mark options={scale = 1.0pt,color=blue,fill=blue}]coordinates{
			(408,0.964)(800,0.901)(1400,0.843)(2500,0.681)(4375,0.649)(8554,0.453)(12749,0.401)(16956,0.383)(20898,0.373)(29299,0.364)(45266,0.340)(57120,0.325)(84353,0.302)(119256,0.274)(149776,0.282)
		};\label{bh2}
		\addplot+[mark=x,only marks,mark options={thick,scale = 1.5pt,color=red,fill=red}]coordinates{
			(408,0.951)(800,0.961)(1397,0.951)(2512,0.910)(4396,0.896)(8580,0.802)(12802,0.701)(17006,0.597)(20898,0.499)(25062,0.411)(29274,0.291)(33143,0.268)(36975,0.244)(49177,0.215)(61025,0.176)(76486,0.153)(92079,0.123)(107337,0.120)(150480,0.112)
		};\label{bh3}
		\draw[dashed,thick] (\pgfkeysvalueof{/pgfplots/xmin},50) -- (\pgfkeysvalueof{/pgfplots/xmax},50);
		\addlegendentry{$B_h=1$}
		\addlegendentry{$B_h=2$}
		\addlegendentry{$B_h=3$}
		\end{axis};

		\end{tikzpicture}
		
		
	\caption{Entropy of the extracted \acp{FSC} from an \ac{RNN} as it is trained with more samples. For each sequence we fix the discretization, ignore the inner loop and add more samples guided by the counterexamples. Note: the behavior at the right edge of the figure is due to the fact that this represents the entropy of the entire \ac{FSC}, which will differ from the entropy over the components of the counterexamples.}
	\label{fig:plot}
\end{figure}
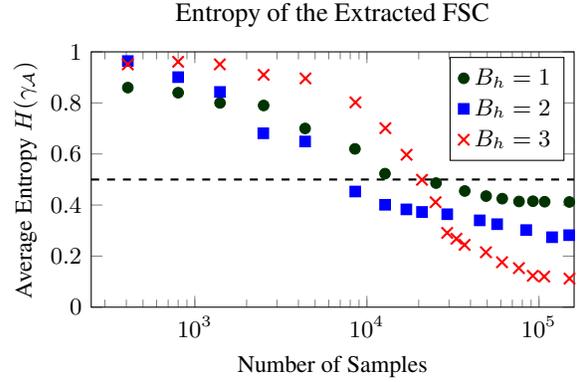
Our goal is to determine whether a policy network requires more training data or not.
Existing approaches in supervised learning methods leverage benchmark comparisons between a train-test set using a loss function~\cite{baum1988supervised}.
Loss visualization, proposed by~\cite{goodfellow2014qualitatively,yu2019interpreting} provides a set of analytical tools to show model convergence.
However, such approaches are generally more suited to classes of continuous functions than the discrete representations we seek.
More importantly, we want to leverage the information we gain from employing a model-based approach. 

%
\paragraph{Counterexamples.}
We first determine a set of states that are critical for the specification under the current strategy. 
Consider the sequences of memory nodes and observations  $(n_0,\obs_0)\xrightarrow{a_0}\cdots \xrightarrow{a_{t-1}}(n_t,\obs_t)$ from the POMDP $\pomdp$ under the \ac{FSC} $\fsc_{\hat{\gamma}}$.
For each of these sequences, we collect the states $s\in S$ underlying the observations, e.g., $\ObsFun(s)=\obs_i$ for $0\leq i\leq t$. 
As we know the probability or expected reward for these states to satisfy the specification from previous model checking, we can now directly assess their criticality regarding the specification.
We collect all pairs of memory nodes and states from $N\times S$ that contain critical states and build the set $\mathit{Crit^\pomdp_{\fsc_{\hat{\gamma}}}}\subseteq N\times S$ that serves us as a counterexample. 
These pairs carry the joint information of critical states and memory nodes from the policy applied to the MC and may be formalized using a so-called product construction.
%
%
%

\paragraph{Entropy measure.}
The average entropy across the distributions over actions at the choices induced by $\mathit{Crit^\pomdp_{\fsc_{\hat{\gamma}}}}$ is our measure of choice to determine the level of training for the policy network. 
Put differently, for each pair $(n,s)\in\mathit{Crit^\pomdp_{\fsc_{\hat{\gamma}}}}$, we collect the distribution $\distFunc\in\Distr(\Act)$ over actions that $\fsc_{\hat{\gamma}}$ returns for the observation $\ObsFun(s)$ when it is in memory node $n$.
Then, we define the \emph{evaluation function} $H$ using the entropy $\mathcal{H}(\distFunc)$ of the distribution $\distFunc$:
\begin{align}
	H\colon Crit^\pomdp_{\fsc_{\hat{\gamma}}}\rightarrow [0,1] \text{ with } H(n,s)=\mathcal{H}(\distFunc)
\end{align}
For high values of $H$, the distribution is uniform across all actions and the associated policy network is likely extrapolating from unseen inputs.

In Fig.~\ref{fig:plot}, we observe that when there are fewer samples and higher discretization, the extracted \ac{FSC} tends to perform arbitrarily.
We define the function $H$ for the full set $\mathit{Crit^\pomdp_{\fsc_{\hat{\gamma}}}}$:
\begin{align}
H(\mathit{Crit^\pomdp_{\fsc_{\hat{\gamma}}}}) &= \frac{1}{|Crit^\pomdp_{\fsc_{\hat{\gamma}}}|} \sum_{(n,s) \in Crit^\pomdp_{\fsc_{\hat{\gamma}}}} H(n,s) \label{eq:ent}
\end{align}
We compare the average entropy over all components of the counterexample against a threshold $\eta \in [0,1]$, that is, if $H(\mathit{Crit^\pomdp_{\fsc_{\hat{\gamma}}}}) > \eta$, we will provide more data.

\addtocounter{example}{-1}
\begin{example}[cont.]
	Under the working example, the policy $\fsc_1$ was the 1-FSC with $p=1$ (Fig.~\ref{fig:1FSC}), which produces two counterexample memory and state pairs: $Crit^\pomdp_{\fsc_1}=\{(0,s_0),(0,s_1)\}$.
	The procedure would then examine the policy's average entropy at these critical components $(n,s) \in Crit^\pomdp_{\fsc_1}$, which in this trivial example is given by $H(Crit^\pomdp_{\fsc_1})=-p \log_2(p) - (1-p) \log_2(1-p)=0$ from (\ref{eq:ent}).
	The average entropy is below a prescribed threshold, $\eta=0.5$, and thus we increase the number of memory nodes, which results in the satisfying \ac{FSC} $\fsc_2$ in Fig.~\ref{fig:2FSC}.
\end{example}

\paragraph{Collecting new training data.}
To perform retraining we take a policy for the underlying MDP $\mdp$ that satisfies $\varphi$, and we use that policy to generate observation-action sequences, which are initialized at the states $s$ in the critical set $Crit^\pomdp_{\fsc_{\hat{\gamma}}}$.
These executions, with observations as inputs and actions as labels, form a batch for retraining the \ac{RNN} policy network $\hat{\osched}$.

\section{Experiments}

We evaluate the proposed verification and synthesis approach by comparing to a series of benchmark examples that are subject to either LTL or expected cost specifications.
For both sets we compare to two synthesis tools: \tool{PRISM-POMDP} \cite{NPZ17} and \tool{SolvePOMDP}\cite{walraven2017accelerated} from the respective formal methods and planning communities.
We further compare to another RNN-based synthesis procedure with a handcrafted memory structure for \acp{FSC} \cite{carr2019counterexample}.
For proper comparison with the synthesis tools we adopt the experiment setup of \cite{carr2019counterexample}, whereby we always iterate for 10 instances of retraining the \ac{RNN} from counterexample data.
Similarly, the proposed approach is not guaranteed to reach the optimum, but shall rather improve as far as possible within 10 iterations.

\paragraph{Implementation and Setup.} We provide a \tool{Python} toolchain that employs the probabilistic model checker {PRISM}. 
We train and encode the \ac{RNN} policy networks $\RNNfun$ using the deep learning library \tool{Keras}. 
All experiments are run on a 2.3 GHz machine with a 12~GB memory limit and a maximum computation time of $10^5$ seconds. 

\paragraph{Settings.}
We analyze the method on three different settings: Maze($c$), Grid($c$) and Navigation($c$), for detailed descriptions of these examples see \cite{NPZ17} and \cite{carr2019counterexample}, respectively.
In each of these settings the policies have an action space of the cardinal directions navigating through a grid-based environment.
For the former two examples, we attempt to synthesize a policy that minimizes an \emph{expected cost} subject to reaching a goal state $a$: $\Ex_{\min}^\pomdp(\Ever a)$.
In the latter example, we seek a policy that maximizes the probability of satisfying an \emph{\ac{LTL} specification}, in particular avoiding obstacles $X$, both static and  randomly moving, until reaching a target area $a$: $\p_{\max}^\pomdp(\neg X \Until a)$.

\paragraph{Discussion.}
The results are shown in Table~\ref{tab:Res}.
The proposed extraction approach scales to significantly larger examples than both state-of-the-art \ac{POMDP} solvers which compute near-optimal policies.
While the handcrafted approach scales equally well, the extraction method produces higher-quality policies - within $2\%$ of the optimum.
That effect is due to our automatic extraction of suitable \acp{FSC}.
Note that an optimal policy for Maze(1) can be expressed using 2 memory states. The \ac{FSC} structure employed by the handcrafted method uses this structure and consequently, for the small Maze environments, the handcrafted method synthesizes higher values.
Yet, with larger environments the fixed memory structure produces poor policies as more memory states are beneficial to account for the past behavior.

\section{Conclusion}
We introduced a novel synthesis procedure for extracting and verifying \ac{RNN}-based policies.
In comparison to other approaches to verify \acp{RNN}, we take a model-based approach and provide guarantees for concrete applications modeled by \acp{POMDP}.
Based on the verification results, we propose a way to either improve the extracted policy or the \ac{RNN} itself.
Our results demonstrate the effectiveness of the approach and that we are  competitive to state-of-the-art synthesis tools.


\bibliographystyle{named}
\bibliography{literature}

\end{document}